\documentclass[10pt,twocolumn,letterpaper]{article}

\usepackage{cvpr}              

\usepackage{graphicx}
\usepackage{amsmath}
\usepackage{amssymb}
\usepackage{booktabs}
\usepackage{multirow}
\newcommand*\samethanks[1][\value{footnote}]{\footnotemark[#1]}

\usepackage[pagebackref,breaklinks,colorlinks]{hyperref}

\usepackage[capitalize]{cleveref}
\crefname{section}{Sec.}{Secs.}
\Crefname{section}{Section}{Sections}
\Crefname{table}{Table}{Tables}
\crefname{table}{Tab.}{Tabs.}

\def\cvprPaperID{360} 
\def\confName{CVPR}
\def\confYear{2023}

\begin{document}

\title{Ultra-High Resolution Segmentation with Ultra-Rich Context: A Novel Benchmark} 

\author{
    Deyi Ji\textsuperscript{\rm1,}\textsuperscript{\rm2} ~~
    Feng Zhao\textsuperscript{\rm1}\thanks{Corresponding Authors.} ~~
    Hongtao Lu\textsuperscript{\rm3,}\textsuperscript{\rm4}\samethanks[1] ~~
    Mingyuan Tao\textsuperscript{\rm2} ~~
    Jieping Ye\textsuperscript{\rm2}  \\
    \textsuperscript{\rm1}University of Science and Technology of China ~~~ \textsuperscript{\rm2}Alibaba Group ~~~ \\
    \textsuperscript{\rm3}Department of Computer Science and Engineering,  Shanghai Jiao Tong University \\
    \textsuperscript{\rm4}MOE Key Lab of Artificial Intelligence, AI Institute, Shanghai Jiao Tong University \\
    {\tt\small jideyi@mail.ustc.edu.cn} ~
    {\tt\small fzhao956@ustc.edu.cn} ~
    {\tt\small htlu@sjtu.edu.cn} ~ \\
    {\tt\small \{juchen.tmy, yejieping.ye\}@alibaba-inc.com} \\
}

\maketitle

\begin{abstract}
   With the increasing interest and rapid development of methods for Ultra-High Resolution (UHR) segmentation, a large-scale benchmark covering a wide range of scenes with full fine-grained dense  annotations is urgently needed to facilitate the field. To this end, the URUR dataset is introduced, in the meaning of \textbf{U}ltra-High \textbf{R}esolution dataset with \textbf{U}ltra-\textbf{R}ich Context.
   As the name suggests, URUR contains amounts of images with high enough resolution (3,008 images of size 5,120$\times$5,120), a wide range of complex scenes (from 63 cities), rich-enough context (1 million instances with 8 categories) and fine-grained annotations (about 80 billion manually annotated pixels), which is far superior to all the existing UHR datasets including DeepGlobe, Inria Aerial, UDD, etc.. Moreover, we also propose WSDNet, a more efficient and effective framework for UHR segmentation especially with ultra-rich context. Specifically, multi-level Discrete Wavelet Transform (DWT) is naturally integrated to release computation burden while preserve more spatial details, along with a Wavelet Smooth Loss (WSL) to reconstruct original structured context and texture with a smooth constrain. Experiments on several UHR datasets demonstrate its state-of-the-art performance. The dataset is available at https://github.com/jankyee/URUR.
   
\end{abstract}

\section{Introduction}
\label{sec:intro}

Benefited from the advancement of photography and sensor technologies, the accessibility and analysis of ultra-high resolution (UHR) images has opened new horizons for the computer vision community, playing an increasingly important role in a wide range of 
applications, including but not limited to disaster control, environmental monitoring, land resource protection and  urban planning. The focus of this paper is on semantic segmentation for UHR images.

\begin{figure*}
    \centering
    \includegraphics[width=1\linewidth]{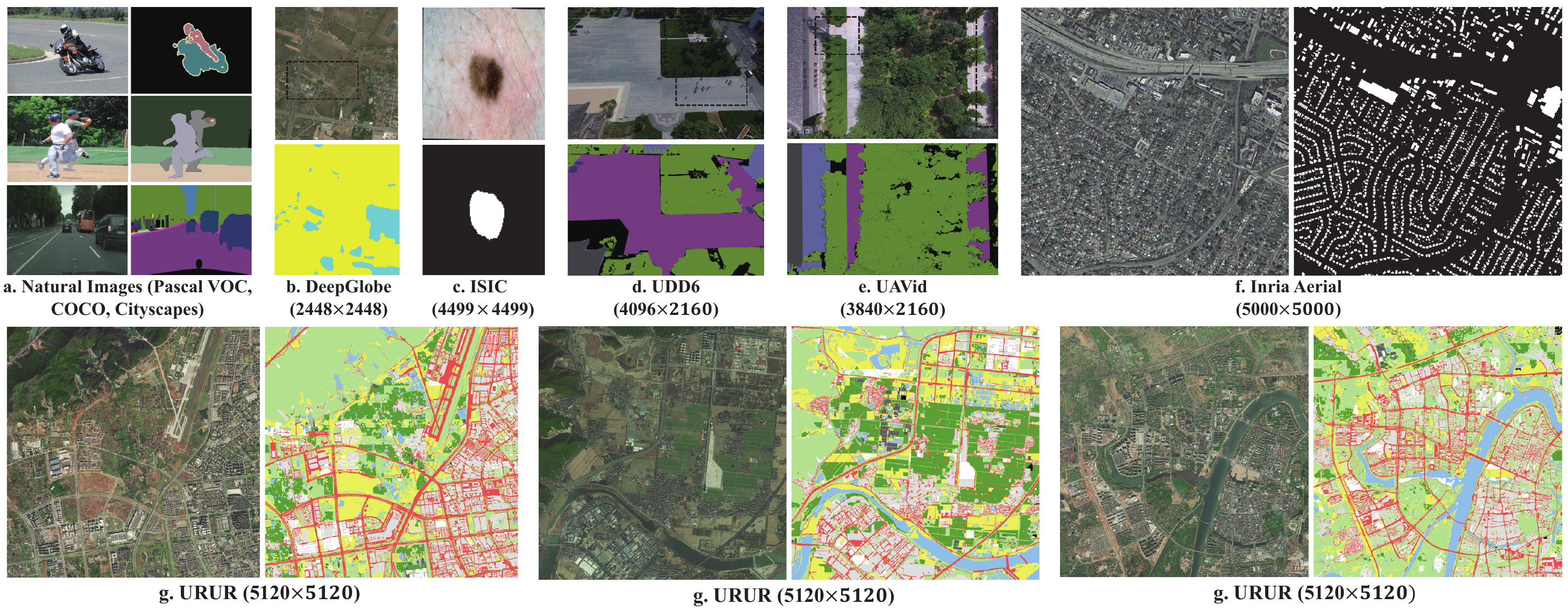}
    \caption{The comparison between natural datasets (Pascal VOC\cite{pascal}, COCO\cite{coco}, Cityscapes\cite{cityscapes}),  and representative UHR datasets (DeepGlobe\cite{deepglobe}, ISIC\cite{isic}, UDD6\cite{udd}, UAVid\cite{uavid}, Inria Aerial\cite{inria} and URUR). As shown that UHR images (from b to g) cover a larger filed of view and contain more regions with very large contrast in both scale and shape, than natural images (a). Existing UHR datasets either adopt coarse annotations (b, d, e) or only annotate one category (c, f). The proposed URUR dataset (h) utilizes fine-grained dense annotations for whole 8 categories.}
    \label{fig_intro}
\end{figure*}

The most commonly-used datasets in existing UHR segmentation methods include DeepGlobe \cite{deepglobe}, Inria Aerial \cite{inria} and Citysacpes\cite{cityscapes}. According the definition of UHR medias\cite{ascher2007filmmaker,glnet}, an image with at least 2048$\times$1080 (2.2M) pixels are regarded as 2K high resolution media. An image with at least 3,840$\times$1,080 (4.1M) pixels reaches the bare minimum bar of 4K resolution, and 4K ultra-high definition media usually refers to a minimum resolution of 3,840$\times$2,160 (8.3M). However, except for Inria Aeral which reaches to 5,000$\times$5,000 pixels, the average resolution of all other two datasets are below 2,500$\times$2,500 (6.2M), thus actually they are not strictly UHR medias. Besides, DeepGlobe also adopts coarse annotations that result in numbers of noises. Although the utra-high resolution, Inria Aerial contains only 180 images in limited scenes, and only annotates one category of building, which is not sufficient to fully verify the performance of UHR segmentation methods and limits the development of the community. Therefore, a novel large-scale benchmark dataset covering a wide range of scenes with full fine-grained dense annotations is urgently needed to facilitate the field. To this end, the URUR dataset is proposed in the paper, in this meaning of \textbf{U}ltra-High \textbf{R}esolution dataset with \textbf{U}ltra-\textbf{R}ich Context. Firstly for the resolution, URUR contains 3,008 UHR images of size  5,120$\times$5,120 (up to 26M), coming from  a wide range of complex scenes in 63 cities. For annotations, there are 80 billion manually annotated pixels, including 2 million fine-grained instances with 8 categories, which is of ultra-high context and far superior to all the existing UHR datasets. Visualization samples and detailed statistics are revealed in Figure \ref{fig_intro} and Section \ref{sec_urur}.

In order to balance the memory occupation and accuracy when the image resolution grows to ultra-high, earlier works for UHR segmentation utilize a two-branch global-local collaborative network to preserve both global and local information, taking the globally down-sampled image and locally cropped patches as inputs respectively. The representative works include GLNet \cite{glnet} and FCtL \cite{fctl}. However, this type of framework requires multiple predictions on the patches thus the overall inference speed is very slow. To further achieve a better balance among accuracy, memory and inference speed, ISDNet \cite{isdnet} is proposed to integrate shallow and deep networks for efficient segmentation. The shallow branch has fewer layers and faster inference speed, its input does not need any downsampling or cropping. For the deep branch, the input image is directly down-sampled to ensure high inference speed. Then a heavy relation-aware feature (RAF) module is utilized to exploit the relationship between the shallow and deep feature. In this paper, we propose WSDNet, the evolution of ISDNet, to formulate a more efficient and effective framework for UHR segmentation. Specifically, multi-level Discrete Wavelet Transform (DWT) and Inverse Discrete Wavelet Transform (IWT) are naturally integrated to release computation burden while preserve more spatial details in the deep branch, thus RAF can be removed for higher inference speed. The Wavelet Smooth Loss (WSL) is also designed to reconstruct original structured context and texture distribution with the smooth constrain in frequency domain.

Overall, the contributions of this paper are summarized as follows:

\begin{itemize}
    \item We introduce the URUR dataset, a novel large-scale dataset covering a wide range of scenes with full fine-grained dense annotations, which is superior to all the exiting UHR datastes to our knowledge.

    \item WSDNet is proposed to preserve more spatial details with multi-level DWT-IWT, and a Wavelet Smooth Loss is presented to reconstruct original structured context and texture distribution with the smooth constrain in frequency domain.
    
    \item Statistics and experiments demonstrate the superiority of URUR and WSDNet. WSDNet achieves  state-of-the-art balance among accuracy, memory and inference speed on several UHR datasets.
\end{itemize}

\section{Related Work}

\subsection{Generic Semantic Segmentation}
With the rapid development of deep learning \cite{goodfellow2016deep, zhang2015cross, ji2019end, zhang2018context, feng2018challenges}, semantic segmentation have achieved remarkable progress. 
Most of generic semantic segmentation models are based on and aim to improve fully convolutional networks (FCN)\cite{fcn}.
They rely on large receptive field and fine-grained deep features \cite{pspnet, deeplabv3+, ocrnet, segformer, setr, stlnet} or graph modules \cite{cdgc, zhou2020graph, cagcn, wu2019simplifying, ipgn}, which are not appropriate to directly apply to UHR images. Real-time segmentors are proposed to balance the computation cost and performance \cite{icnet, bisenetv2, stdc}. 
BiseNetV2\cite{bisenetv2} achieved considerable performance benefited from its specially designed architectures(bilateral aggregation) and training strategies(booster training). 
However these methods usually rely on small receptive fields and feature channel cutting techniques, which sacrifices the feature view. In addition, knowledge distillation frameworks are also utilized to produce efficient yet high-performance segmentation models \cite{skd, sstkd}.

\subsection{UHR Semantic Segmentation}
Many methods have been especially presented for UHR semantic segmentation\cite{cascadepsp, glnet, fctl, gpwformer, isdnet}.
CascadePSP\cite{cascadepsp} proposed to improve the coarse segmentation results with a pre-trained model to generate high-quality results. GLNet\cite{glnet} firstly incorporated both global and local information deeply in a two-stream branch manner. Based on GLNet, FCtL \cite{fctl} further exploited a squeeze-and-split structure to fuse multi-scale features information. For the sake of higher inference speed, ISDNet \cite{isdnet} directly processed the full-scale and down-sampled inputs by integrating shallow and deep networks, significantly accelerating the inference speed.

\section{URUR Dataset} \label{sec_urur}

The proposed URUR dataset is far superior to all the existing UHR datasets including DeepGlobe, Inria Aerial, UDD, etc., in terms of both quantity, context richness and annotation quality. In this section, we illustrate the processes of dataset construction and analyze them through a variety of informative statistics, as well as give detailed measures to protect privacy.

\begin{table*}[t]
\centering
\scalebox{0.87}{
\begin{tabular}{c|cc|ccccc|cc|cc}
\toprule
\multicolumn{1}{c|}{\multirow{2}{*}{\begin{tabular}[c]{@{}c@{}} \\ \textbf{UHR} \\ \textbf{Dataset} \end{tabular}}} & 
\multicolumn{2}{c|}{\textbf{Image Statistics}} & 
\multicolumn{5}{c|}{\textbf{Overall Annotated Statistics}} & 
\multicolumn{2}{c|}{\textbf{Per Annotated Statistics}}& 
\multicolumn{2}{c}{\textbf{Scene Complexity}} \\ \cmidrule{2-12} 

\multicolumn{1}{c|}{}  & Img. & 

Resolution &
Type &
Pixels &
Density &
Cls. &
Inst. &
\begin{tabular}[c]{@{}c@{}} Ave. Cls. per \\  Img./Region \end{tabular} & 
\begin{tabular}[c]{@{}c@{}} Ave. Inst. per \\ Img./Region \end{tabular} & 
Cities & 
\begin{tabular}[c]{@{}c@{}} Context \end{tabular}  \\ \midrule

\multicolumn{1}{c|}{DeepGlobe}               & 803    & 2448$\times$2448  & coarse     & 4812M & 1.0 & 8                     & 21K                    & 3.9/1.8                                    &    17/4.9                            & 3      &  0.398   \\
\multicolumn{1}{c|}{Inria Aerial}            & 180    & 5000$\times$5000 & fine       &   710M & 0.16 & 2                     & 138K                    & 2/0.8                                    &  766/302                         & 10        & 0.367   \\
\multicolumn{1}{c|}{ISIC$^*$}                    & 2596   & 6682$\times$4401 & coarse     &    247M     &   0.01       & 2                     & 2.6K                  & 2/0.2                                    &  1/0.2                             & -      &  0.087    \\
\multicolumn{1}{c|}{ERM-PAIW}                & 33    & 4795$\times$3014 & fine       &  71M  & 0.15 & 2                     & 0.3K                    & 2/0.6                                    &    1/0.6                        & 11      &  0.277  \\
\multicolumn{1}{c|}{UDD6}                     & 141    & 4096$\times$2160 & coarse  &   1250M  &   1.0   & 6                     & 21K                    & 5.8/3.4                                  &   99/42                                 & 4      &   0.471   \\
\multicolumn{1}{c|}{UAVid}                   & 140    & 3840$\times$2160 & coarse      &   1001M  & 1.0 & 8                     & 22K                    & 6.6/4.1                                    &  93/54                                & 1      &  0.459    \\ \midrule
\multicolumn{1}{c|}{URUR}                    & 3008   & 5120$\times$5120 & fine       &   78852M  & 1.0  & 8                     & 1140K                    & 7.2/5.6                                   &         379/201                     & 63     &  0.883      \\ \bottomrule
\end{tabular}}
\caption{The detailed statistics comparison between URUR and existing UHR datasets, including DeepGlobe\cite{deepglobe}, Inria Aerial\cite{inria}, ISIC\cite{isic}, ERM-PAIW\cite{erm}, UDD6\cite{udd} and UAVid\cite{uavid}. As shown that URUR  is  far  superior  to  all  of them in terms of both quantity, annotation quality, context richness and scene complexity. ``Img.", ``Cls.", ``Inst.", ``Ave." denote ``Image", ``Class/Category", ``Instance", ``Average" respectively. For UDD6 and UAVid, the testing sets are not included since the annotations have not been open sourced. The resolution of images in ISIC is diverse and the largest is up to 6682$\times$4401. Instances that are too small are not considered.}
\label{table_sta}
\end{table*}

\subsection{Dataset Summary}

The proposed URUR dataset contains 3,008 UHR images with size of 5,012$\times$5,012, captured from 63 cities. The training, validation and testing set include 2,157, 280 and 571 UHR images respectively, with the approximate ratio of 7:1:2. All the images are exhaustively manually annotated with fine-grained pixel-level categories, including 8 classes of ``building", ``farmland", ``greenhouse", ``woodland", ``bareland", ``water", ``road"  and ``others".  Sample images are shown in Figure \ref{fig_intro} (h). The number of images and annotations in the dataset is still growing.

\subsection{Data Collection and Pre-processing}

 The dataset is collected by several high-quality satellite image data sources for public use. This results in data from 63 cities which we then select about 20 scenes manually in each city, based on following standards: 

\begin{itemize}
    \item Low Ambiguity: The objects in the selected scenes should not have much obvious semantic ambiguity in appearance.
    \item High Diversity: Scenes with diverse types of categories, instances, times and weather should be more appropriate and meaningful in our task.
    \item Privacy Protection: No information in the scenario should reveal anything about privacy, such as person, store name, etc.
\end{itemize}
Therefore, the dataset has a high variation in camera viewpoint, illumination and scenario type. In addition, in order to enhance the diversity and richness of the dataset, multiple granular perspectives are set and collected for each scenario. As a result,  we totally collect 752 images with size 10,240$\times$10,240, which are then divided to 3,008 images with size 5,120$\times$5,120.

\subsection{Efficient Annotation}

Compared to natural images, annotating the UHR images is always a more tough job, since the objects to be labeled grow quadratically as the image resolution increases. This is why existing UHR datasets usually exploit coarse-grained annotations or annotate only one major category. In contrast, we are intended to adopt more fine-grained annotations for the whole categories in the proposed URUR dataset. Figure \ref{fig_intro} shows an intuitive comparison and more details about dataset statistics will be presented on Section \ref{data_sta}. As seen that the UHR datasets including DeepGlobe, Inria Aerial and URUR obviously contain more objects and instances than natural ones, such as  Pascal VOC and COCO, while the objects are also smaller in scale. Moreover, one or more class pairs are often spatially mixed together, bringing great troubles to carefully distinguish them during annotation process. By contrast, URUR also contains more objects and richer context than other UHR datasets. In conclusion, the main challenge and time-consuming part of annotating fine-grained UHR images are not only reflected in the amounts of objects to be annotated  caused by the excessively ultra-high image resolution, but also in the many chain problems caused by the ultra-rich image context among objects with drastically changing scales. 

For both efficient and accurate annotation, each original UHR image with size of 5,120$\times$5,120 is firstly cropped evenly into multiple patches with size of 1,000$\times$1,000. We let the annotators annotate these image patches separately, after that their results are correspondingly merged to get the final annotations relative to the original UHR images. In this way, we ensure that each annotator only focus on a smaller image patch, which facilitates the annotation process and improves the accuracy of the annotation results. During cropping, neighboring patches have 120$\times$1,000 pixels overlap region to guarantee the consistency of annotation results and avoid boundary vanishing. 
In order to further save manpower and speed up the whole process, a ISDNet model is trained  with the early manually annotated images and used to generate segmentation masks on the rest images. As a reference, annotators adjust the masks with the help of annotation tools developed by us.

\subsection{Dataset Statistics} \label{data_sta}

Table \ref{table_sta} shows the detailed statistics comparison between the proposed URUR dataset and exiting several main UHR datasets, including DeepGlobe \cite{deepglobe}, Inria Aerial \cite{isic}, ISIC \cite{isic}, ERM-PAIM\cite{erm}, UDD \cite{udd} and UAVid \cite{uavid}. First of all, for the most fundamental image statistics, URUR consists of 3,008 images with size of 5,120$\times$5,120 and outperforms all other datasets on both image number and resolution. In concrete, except for ISIC and DeepGlobe, the image number of all other datasets are below 200. DeepGlobe contains 803 images but the resolution is only 2,448$\times$2,448 (5.9M), which does not even reach the minimum threshold (8.3M) of UHR medias (illustrated in Section \ref{sec:intro}). For overall annotation, limited by manpower, the annotation paradigms of existing datasets are divided into two types: (1) using coarse annotation, or (2) only annotating one category. The first type includes DeepGlobe, UDD6 and UAVid. As samples shown in Figure \ref{fig_intro} (b, d, e), a large area of land containing many farmlands and buildings has been directly annotated as bareland for simplify in DeepGlobe. The cars, persons and trees are directly roughly painted in UDD6 and UAVid. The second type includes Inria Aerial and ERM-PAIM, they adopt a fine-grained annotations but only annotate one category of buildings and roads respectively. ISIC is a medicine dataset about lesion segmentation. Although it has up to 2,596 images, there is only one category of lesion area being roughly annotated. By contrast, URUR totally annotates 78,852 million pixels, with 100\% annotation density on 8 categories, and the total number of annotated instances are up to 2,058 thousand, which is far superior to all other datasets. More details about per image annotation statistics are also revealed. We count the average number of categories and instances per image, which can reflect the context richness and scene complexity in some degree. For a closer observation, we also randomly sample some regions and count the average categories and instances in them. As found in Table \ref{table_sta}, although both DeepGlobe, UDD6 and UAVid contain multiple categories, their average category per image/region is very low because of the coarse annotations and relative-simple scenes. By contrast, URUR consists of high-density categories and instances in each image with complex scenes. The other meta information is also provided, such as number of cities for data collection.

Finally, we design a quantitative measure metric, namely Scene Context Richness, to compare the overall scene complexity in datasets. Formally, it is defined as follows,
\begin{equation}
    R = -\sum_{c}^C(O_c)^{\frac{1}{q}}\cdot p_c \cdot log(p_c)
\end{equation}
\noindent where $R$ is the context richness, $C$ is the number of categories, $O_c$ is the average number of object instances per region for category $c$, $p_c$ is the average probability of category $c$ per region. Thus we can see that when the dataset contains more object instances and more diverse categories in each region, its overall context is richer thus scene complexity is higher. $q$ is the temperature parameter to adjust the weight of instance number and set to 2 in our experiments. We randomly select some regions for all the datasets and calculate $R$, results show that URUR has the highest scene complexity ($R=0.883$) and ultra-rich context.

\begin{figure*}
    \centering
    \includegraphics[width=1\linewidth]{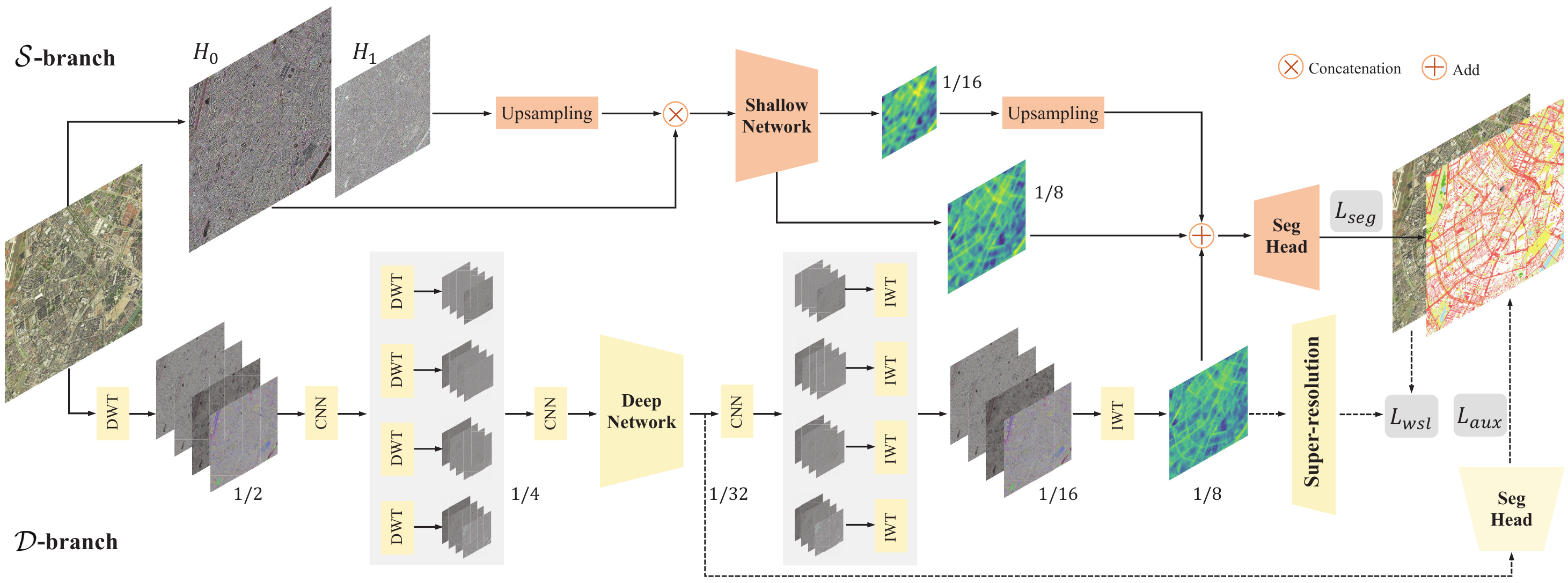}
    \caption{The overview of the proposed WSDNet for UHR segmentation, which consists of a deep branch $\mathcal{D}$ (the lower branch) and a shallow branch $\mathcal{S}$ (the upper branch). In $\mathcal{S}$, the input images is decomposed into two subbands with Laplacian pyramid, which are then concatenated and fed into a shallow network to extract full-scale spatial details. In $\mathcal{D}$, the input image is down-sampled with two-level Discrete Wavelet Transform (DWT) and then fed into the deep network to harvest high-level category-wise context. Next the output with scale $\frac{1}{32}$ of the original input is upsampled to $\frac{1}{8}$ with two-level Invert Discrete Wavelet Transform (IWT). Finally the two branches are fused with multi-scale features and optimized with the base cross-entropy losses $\mathcal{L}_{seg}$, auxiliary loss $\mathcal{L}_{aux}$, as well as a Wavelet Smooth Loss (WSL) to reconstruct the original input with the help of a super-resolution head. The modules within dot lines are removed during inference.}
    \label{fig_method}
\end{figure*}

\subsection{Privacy Protection Statement}

For the most important thing, our dataset is only used for academic purposes to drive the development of UHR image analysis techniques. We have fully considered all the possibilities to avoid anything about privacy issues in the dataset collection stage. The source of dataset comes from satellites for public use and is not related to any sensitive information. Annotators are also asked to filter and discard the potentially sensitive information. Specifically, we ask annotators to cover up or discard any sensitive information in a scene, including time and address watermarks in videos, phone numbers, and addresses on the shops or walls. The primary purpose of this paper is to facilitate the development of this field to the community better. We try to provide a more large-scale, fine-grained and challenging dataset for future researches. All researchers who ask for the dataset should follow the Data Usage Protocol under the legal protection provided by us \cite{wang2021learning}.

\section{WSDNet}

\subsection{Network Architecture}

As shown in Figure \ref{fig_method}, WSDNet consists of a deep branch $\mathcal{D}$ and a shallow branch $\mathcal{S}$. 
$\mathcal{S}$ contains fewer layers without any down-sampling or cropping operations on input UHR image to harvest all spatial details while preserving high inference speed. Following ISDNet\cite{isdnet}, the original input RGB image $I$ is replaced with high-frequency residuals $\{H\}^n_{i=0}$:
\begin{equation}
    H_i = g_i(I) - U(g_{i+1}(I))
\end{equation}
\noindent where $g(\cdot)$ denotes Gaussian blur, $U(\cdot)$ denotes the upsampling operation. The original outputs are two feature maps with $\frac{1}{8}$ and $\frac{1}{16}$ of the original image, and the $\frac{1}{16}$ feature map is then up-sampled to add to $\frac{1}{8}$ feature map for final output.

$\mathcal{D}$ is a deep network responsible for learning category-wise context, and input the $\frac{1}{4}$ down-sampled UHR image for faster inference speed and lower memory occupation. Instead of naive down-sampling in ISDNet, we are intended to exploit an invertible downsampling operation to preserve the original image details with less information loss,
and wavelet transform (DWT) is considered. Wavelet transform is a fundamental time-frequency analysis method that decomposes input signals step by step into different frequency subbands to address the aliasing problem. 
In particular, Discrete Wavelet Transform (DWT) \cite{wavelet_cnn} enables invertible down-sampling by transforming input image $I$ into four discrete wavelet subbands $I_1, I_2, I_3, I_4$ with four filters ($f_{LL}, f_{LH}, f_{HL}$, $f_{HH}$)

\begin{equation}
    \begin{aligned}
        I_1 = (f_{LL} \otimes I) \downarrow 2, \ & I_2 = (f_{LH} \otimes I) \downarrow 2 \\
        I_3 = (f_{HL} \otimes I) \downarrow 2, \ & I_4 = (f_{HH} \otimes I) \downarrow 2.
    \end{aligned}
\end{equation}
\noindent where $\otimes$ is the convolution operation. $I_1$ represents all low-frequency information describing the basic object structure at coarse-grained level. $I_2, I_3, I_4$ include high-frequency information retaining the object texture details at fine-grained level \cite{wavevit}.
In this way, various levels of image details are preserved in different subbands of lower resolution without information dropping. Although down-sampling operation is used, due to the good biorthogonal property of DWT, the original image $I$ can be reconstructed by the Inverse Discrete Wavelet Transform (IWT)\cite{wavelet_cnn}, i.e., $I = IWT(I_1,I_2,I_3,I_4)$.  When integrated to CNN, the DWT-IWT paradigm is able to preserve more spatial and frequency information than ordinary downsampling methods. 

The subband images $I_1, I_2, I_3, I_4$ can be further processed with DWT to produce the decomposition results. For two-level DWT, each subband image $I_b (b \in [1, 4])$ is decomposed into four subband images $I_{b,1}, I_{b,2}, I_{b,3}$, $I_{b,4}$. Recursively, the results of  higher levels DWT can be attained. In $\mathcal{D}$, we integrate two-level DWT with CNN block to obtain the $\frac{1}{4}$ down-sampled input image, followed by  the deep network. The output is $\frac{1}{32}$ feature map rich in high-level category-wise context, and then up-sampled to $\frac{1}{8}$ feature map with two-level IWT. In this way, the output of $\mathcal{D}$ has the same size with the output of $\mathcal{S}$, so they can be naturally fused and no extra special fusion module is required, such as the heavy RAF module in ISDNet. This further accelerates the inference and decreases the memory cost.

\subsection{Wavelet Smooth Loss}

In order to further weaken the affect of down-sampled low-resolution input in $\mathcal{D}$, a super-resolution head is added after the $\frac{1}{8}$ output of $\mathcal{D}$ to reconstruct the original input. Instead of an ordinary super-resolution loss that formulates a hard reconstruction constrain, we propose the Wavelet Smooth Loss (WSL) to optimize the reconstruction process with a soft and smooth constrain, by reconstructing the super-resolution output $I^{rec}$ in frequency domain. More comprehensively, we apply $L$-level DWT to $I$ and $I^{rec}$ respectively, and obtain their low- and high-frequency subbands. The L1 regularization, not the L2 regularization, is used to constrain the high-frequency subbands. Because we prefer to align the texture distribution between $I$ and $I^{rec}$, rather than specific frequency values, but gradient of L2 regularization is closely related to the values, while the gradient of L1 regularization is independent. This type of smooth constraint can make the texture distribution of the output from $\mathcal{D}$ consistent with the input, and avoid the over-fitting caused by the exact numerical alignment in L2 regularization.

On the contrary, since low-frequency subbands represent the basic objects structure details, we exploit L2 regularization to make the spatial structured details of the output fit the ones of input as closely as possible, driving $\mathcal{D}$ to preserve more spatial information. Overall, the WSL consists of the above two parts and is formulated as,

\begin{equation}
\begin{aligned}
    \mathcal{L}_{wsl} = \sum_{l=1}^L \sum_{b=1}^{4^{l}}(&\lambda_1 ||I_{l,b;1} - I_{l,b;1}^{rec}||_2 + \\ &\lambda_2 \sum_{i=2}^{4}||I_{l,b;i} - I_{l,b;i}^{rec}||_1).
\end{aligned}
\end{equation}

\noindent where $I_{l,b;1}$ denotes the low-frequency subband after $l$-th DWT, $I_{l,b;i}$ denotes the $i$-th high-frequency subband after $l$-th DWT. $I_{l,b;1}^{rec}$, $I_{l,b;i}^{rec}$ and so on. $\lambda_1, \lambda_2$ are the weights of the low-frequency and high-frequency constrains respectively.

\subsection{Optimization}

In addition, the standard cross-entropy loss is also used for both the final segmentation results ($\mathcal{L}_{seg}$) and the auxiliary segmentation head after $\mathcal{D}$ ($\mathcal{L}_{aux}$). So the overall loss $\mathcal{L}$ is:
\begin{equation}
    \mathcal{L} = \mathcal{L}_{seg} + \lambda_3 \mathcal{L}_{aux} + \mathcal{L}_{wsl}.
    \label{eq_loss}
\end{equation}
\noindent where $\lambda_3$ is the weight of $\mathcal{L}_{aux}$. Noted that both the reconstruction head and segmentation head in $\mathcal{D}$ are only used during training, and will be removed in inference stage, which are indicated with dot lines in Figure \ref{fig_method}.

\section{Experiments}

\subsection{Datasets and Evaluation Metrics}

We perform extensive experiments on  DeepGlobe, Inria Aerial and URUR dataset to validate WSDNet. In addition to URUR, we describe the former two  datasets  as follows.

\noindent\textbf{DeepGlobe}. The DeepGlobe dataset\cite{deepglobe} has 803 UHR images (455, 207 and 142 for training, validation and testing respectively). Each image contains $2448 \times 2448$ pixels and seven classes of landscape regions, where one class called ``unknown" is not considered in the evaluation. 
Following \cite{glnet,fctl}, we split images into training, validation and testing sets with 455, 207, and 142 images respectively.

\noindent\textbf{Inria Aerial}. The Inria Aerial \cite{inria} has 180 UHR images (126, 27 and 27 for training, validation and testing respectively). Each image contains $5000 \times 5000$ pixels and is annotated with a binary mask for building/non-building areas. 
This datasets covers diverse urban landscapes, ranging from dense metropolitan districts to alpine resorts. Unlike DeepGlobe, it splits the training/test sets by city. We follow the protocol as \cite{glnet,fctl} by splitting images into training, validation and testing sets with 126, 27, and 27 images, respectively.

\noindent\textbf{Evaluation Metrics}. Intersection-over-Union (mIoU), F1 score,  Accuracy and Frames-per-second (FPS) are used to study the effectiveness and inference speed.

\subsection{Implementation Details}

We adopt the mmsegmentation\cite{mmseg2020} toolbox as codebase and follow the default augmentations without bells and whistles. $\mathcal{D}$ and $\mathcal{S}$ can be any usual segmentation networks, here we exploit DeepLabV3+\cite{deeplabv3+} with ResNet18 and STDC\cite{stdc} respectively. 
For fairness comparison, we set the same training settings as \cite{isdnet}: SGD with momentum 0.9 for all parameters are used, the initial learning rate is configured as $10^{-3}$ with polynomial decay parameter of 0.9, batch size is 8 and the maximum iteration number are set to 40K, 80K and 160K on DeepGlobe, Inria Aerial and URUR respectively. In Equation \ref{eq_loss}, $\lambda_1=1, \lambda_2=0.8, \lambda_3 = 0.1, L=3$. We use the command line tool “gpustat” to measure the GPU memory. Memory and Frames-per-second (FPS) are measured on a RTX 2080Ti GPU with a batch size of 1, which is also same as \cite{isdnet}.

\subsection{Comparison with State-of-the-arts}

We compare WSDNet with representative generic and UHR segmentation methods. Since most of generic methods have not specially designed for UHR images, there are two inference paradigms: (1) Global Inference: inference model with the down-sampled global images. (2) Local Inference: inference model with the cropped patches by multiple times then merge their results.

\noindent\textbf{DeepGlobe}. Although DeepGlobe is not strictly an UHR dataset, we still follow previous works and use it as a reference to validate the effectiveness of WSDNet. Compared with both generic and UHR models in Table \ref{sota_deepglobe}, WSDNet achieves excellent balance between mIoU, F1, accuracy, memory and FPS. In concrete, due to multiple patch inferences, the overall inference speed of GLNet\cite{glnet} and FCtL\cite{fctl} is very low. Compared with ISDNet, WSDNet removes the heavy RAF module thus the inference speed is further increased from 27.7 to 30.3. Moreover, benefited from the DWT-IWT paradigm and WSL, the performance is also further improved.

\begin{table}[t]
    \centering
    \scalebox{0.95}{\begin{tabular}{l c c c c c}
      \toprule
      \textbf{Generic Models} & 
      \begin{tabular}[c]{@{}c@{}} mIoU \\ (\%)$\uparrow$ \end{tabular} & 
      \begin{tabular}[c]{@{}c@{}} F1 \\ (\%)$\uparrow$ \end{tabular} &  \begin{tabular}[c]{@{}c@{}} Acc \\ (\%)$\uparrow$ \end{tabular} & \begin{tabular}[c]{@{}c@{}} Mem \\ (M)$\downarrow$ \end{tabular} & \begin{tabular}[c]{@{}c@{}} FPS \\$\uparrow$ \end{tabular} \\ \midrule
    \textbf{\textit{Local Inference}} & & & & &  \\
    U-Net\cite{unet} & 37.3  & -   & -  &  949 & 1.26 \\
    
    DeepLabv3+\cite{deeplabv3+} & 63.1 & -  & - & 1279 & 1.60   \\
    FCN-8s\cite{fcn} & 71.8  & 82.6 & 87.6 & 1963 & 4.55 \\
    \midrule
    
    \textbf{\textit{Global Inference}} & & & &  \\
    U-Net\cite{unet} & 38.4  & -   & -  & 5507 & 3.54 \\
    ICNet\cite{icnet}  & 40.2  & - & - & 2557 & 5.3   \\
    PSPNet\cite{pspnet} & 56.6 & - & - & 6289 & 1.0  \\
    DeepLabv3+\cite{deeplabv3+} & 63.5 & -  & - & 3199 & 4.44  \\
    FCN-8s\cite{fcn} & 68.8  & 79.8 & 86.2 & 5227 & 7.91 \\
    BiseNetV1\cite{bisenet} & 53.0 & - & - & 1801 & 14.2 \\
    DANet\cite{danet} & 53.8 & - & - & 6812 & 2.3 \\
    STDC\cite{stdc} & 70.3 & - & - & 2580 & 14.0 \\
    \toprule
    
    \textbf{UHR Models} &  &  &  &  &  \\ \midrule
    CascadePSP\cite{cascadepsp} & 68.5 & 79.7 & 85.6 & 3236 & 0.11  \\
    PPN\cite{ppn} & 71.9 & - & - & 1193 & 12.9 \\
    PointRend\cite{pointrend} & 71.8 & - & - & 1593 & 6.25 \\
    MagNet\cite{magnet} & 72.9 & - & - & 1559 & 0.80 \\
    MagNet-Fast\cite{magnet} & 71.8 & - & - & 1559 & 3.40 \\
    GLNet\cite{glnet}  & 71.6 & 83.2 & 88.0  & 1865 & 0.17\\
    FCtL\cite{fctl}  & 72.8 & 83.8 & 88.3 & 3167 & 0.13 \\ 
    ISDNet\cite{isdnet} & 73.3 & 84.0 & 88.7  & 1948 & 27.7 \\
    \textbf{WSDNet} & \textbf{74.1} & \textbf{85.2} & \textbf{89.1} & 1876 & \textbf{30.3} \\
     
    \bottomrule
    \end{tabular}}
    \caption{Comparison with state-of-the-arts on DeepGlobe \textit{test} set. ``Acc", ``Mem" indicates ``Accuracy", ``Memory" respectively, the same below}
    \label{sota_deepglobe}
\end{table}

\begin{table}[t]
    \centering
    \scalebox{0.95}{\begin{tabular}{l c c c c c}
      \toprule
      \textbf{Generic Models} & 
      \begin{tabular}[c]{@{}c@{}} mIoU \\ (\%)$\uparrow$ \end{tabular} & 
      \begin{tabular}[c]{@{}c@{}} F1 \\ (\%)$\uparrow$ \end{tabular} &  \begin{tabular}[c]{@{}c@{}} Acc \\ (\%)$\uparrow$ \end{tabular} & \begin{tabular}[c]{@{}c@{}} Mem \\ (M)$\downarrow$ \end{tabular} & \begin{tabular}[c]{@{}c@{}} FPS \\$\uparrow$ \end{tabular} \\ \midrule
    
    DeepLabv3+\cite{deeplabv3+} & 55.9 & -  & - & 5122 & 1.67   \\
    FCN-8s\cite{fcn} & 69.1  & 81.7 & 93.6 & 2447 & 1.90\\
    STDC\cite{stdc} & 72.4 & - & - & 7410 & 4.97 \\
    \toprule
    
    \textbf{UHR Models} &  &  &  &  &  \\ \midrule
    CascadePSP\cite{cascadepsp} & 69.4 & 81.8 & 93.2 & 3236 & 0.03  \\
    GLNet\cite{glnet}  & 71.2 & - & -  & 2663 & 0.05\\
    FCtL\cite{fctl}  & 73.7 & 84.1 & 94.6 & 4332 & 0.04 \\
    ISDNet\cite{isdnet} & 74.2 & 84.9 & 95.6  & 4680 & 6.90 \\
    \textbf{WSDNet}  & \textbf{75.2} & \textbf{86.0} & \textbf{96.0} & 4379 & \textbf{7.80} \\
     
    \bottomrule
    \end{tabular}}
    \caption{Comparison with state-of-the-arts on Inria Aerial \textit{test} set}
    \label{sota_inria}
\end{table}

\begin{table}[t]
    \centering
    \scalebox{1}{\begin{tabular}{l c  c c c}
      \toprule
      \textbf{Generic Models} & 
      \begin{tabular}[c]{@{}c@{}} mIoU \\ (\%)$\uparrow$ \end{tabular} & 
      \begin{tabular}[c]{@{}c@{}} Acc \\ (\%)$\uparrow$ \end{tabular} & \begin{tabular}[c]{@{}c@{}} Mem \\ (M)$\downarrow$ \end{tabular} & \begin{tabular}[c]{@{}c@{}} FPS \\$\uparrow$ \end{tabular} \\ \midrule
    
    PSPNet\cite{pspnet} & 32.0   & - & 5482 & 1.86 \\
    DeepLabv3+\cite{deeplabv3+} & 33.1   & - & 5508 & 1.97   \\
    STDC\cite{stdc} & 42.0  & - & 7617 & 4.31 \\
    \toprule
    
    \textbf{UHR Models} &  &    &  &  \\ \midrule
    GLNet\cite{glnet}  & 41.2  & 71.5  & 3063 & 0.04\\
    FCtL\cite{fctl}  & 43.1  & 73.8 & 4508 & 0.03 \\
    ISDNet\cite{isdnet} & 45.8  & 75.6  & 4920 & 6.31 \\
    \textbf{WSDNet}  & \textbf{46.9} &  \textbf{76.8} & 4560 & \textbf{7.13} \\
     
    \bottomrule
    \end{tabular}}
    \caption{Comparison with state-of-the-arts on URUR \textit{test} set}
    \label{sota_urur}
\end{table}

\noindent\textbf{Inria Aerial}. Inria Aerial is an actual UHR dataset with size 5,000$\times$5,000 thus more convincing to prove the superiority. It is only annotated one category of building and Table \ref{sota_inria} shows the comparisons.
WSDNet also achieves the better balance among all metrics.

\noindent\textbf{URUR}.
Due to ultra-high resolution, ultra-rich fine-grained annotations and ultra-diversity of land cover types, URUR is the most challenging UHR datasets so far, compared to all other datasets. As shown in Table \ref{sota_urur}, WSDNet also outperforms existing methods by a very large margin on mIoU, while preserving higher inference speed.

\subsection{Ablation Study}

In all ablation studies, we perform experiments on URUR \textit{test} set to validate the effectiveness of each component.

\begin{table}[t]
    \centering
    \scalebox{1}{\begin{tabular}{l | c c}
      \toprule
      Downsampling & mIoU(\%) & FPS \\ \midrule
    
    uniform downsampling  & 45.1 & 7.65 \\
    multi-level CNN & 45.8 & 5.62 \\ 
    adaptive downsampling & 46.0 & 4.96  \\
    multi-level DWT & 46.9 & 7.13 \\
     
    \bottomrule
    \end{tabular}}
    \caption{Comparison with different down-sampling methods.}
    \label{abl_down-sample}
\end{table}

\begin{table}[t]
    \centering
    \scalebox{1}{\begin{tabular}{l |  c}
      \toprule
      Loss Function & mIoU(\%) \\ \midrule
      baseline($\mathcal{L}_{seg}$ \& $\mathcal{L}_{aux}$) & 45.2 \\
      baseline + $\mathcal{L}_{sr}$ \& $\mathcal{L}_{sd}$ & 45.9 \\
      baseline + $\mathcal{L}_{wsl}$ & 46.9 \\
      \bottomrule
    \end{tabular}}
    \caption{Effectiveness of loss functions.}
    \label{abl_loss}
\end{table}

\subsubsection{Comparison of downsampling methods}

We compare the different types of downsampling methods in Table \ref{abl_down-sample}. The baseline type uses an ordinary uniform downsampling in the form of bilinear interpolation.
We also attempt to realize the downsampling process by a multi-level CNN module with a combination of several convolution and pooling layers. Then an adaptive downsampling method based on deformable convolution \cite{deformable} is also tried. 
Experimental results show DWT-IWT paradigm achieves the best performance on mIoU and considerable inference speed, proving that DWT-IWT paradigm can preserve higher performance  for the input in deep branch than the ordinary down-sampling.

\subsubsection{Effectiveness of WSL}

Table \ref{abl_loss} shows the effectiveness of proposed WSL. The baseline is the  cross-entropy loss $\mathcal{L}_{seg}$ and auxiliary loss $\mathcal{L}_{aux}$. Then we add the ordinary super-resolution loss in \cite{isdnet} and the proposed WSL respectively. Experimental results show WSL achieves highest performance, proving the effectiveness of the smooth constrain in frequency domain.

\subsubsection{Quantitative Results}

\begin{figure}
    \centering
    \includegraphics[width=1\linewidth]{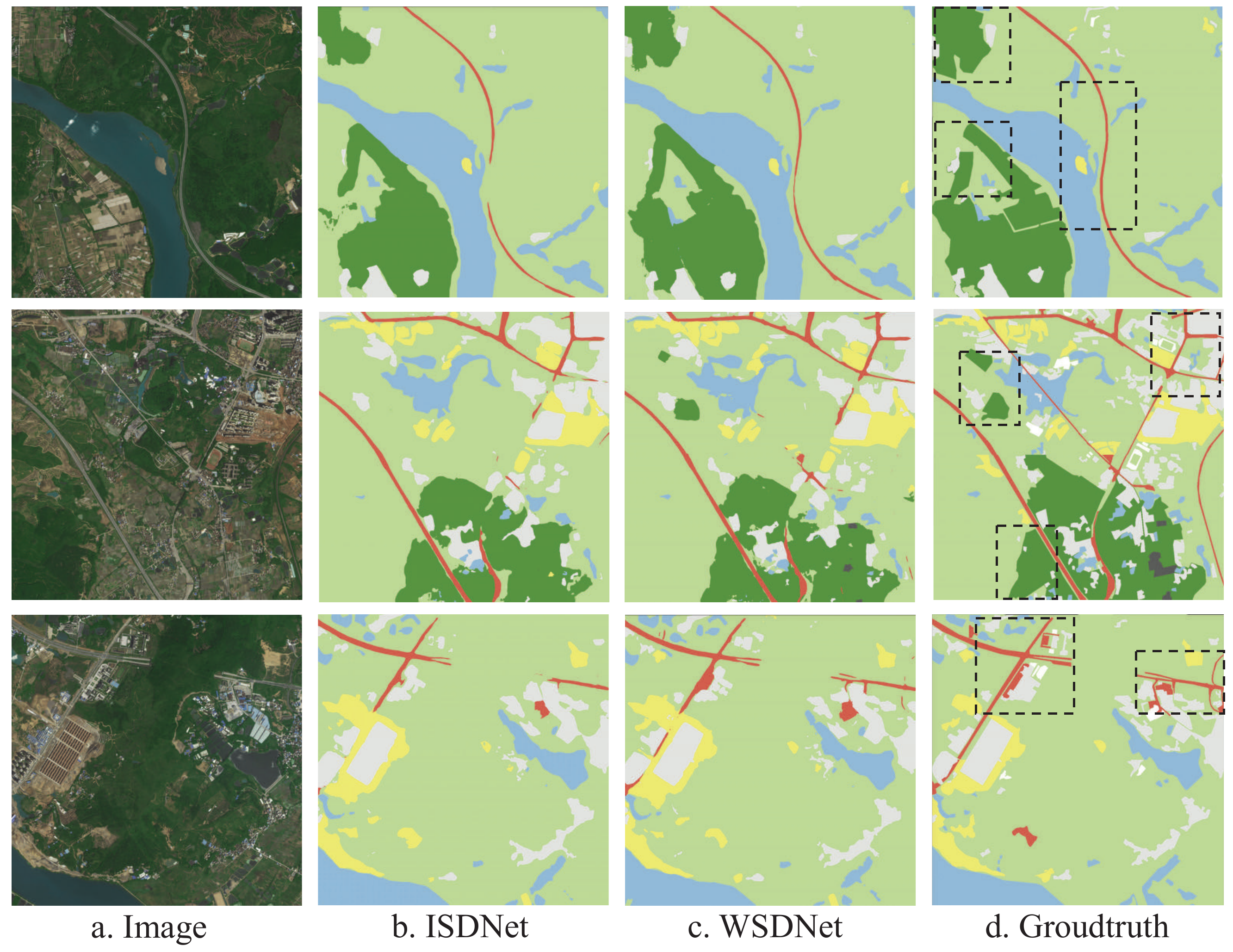}
    \caption{Visual improvements on URUR dataset: (a) part of original UHR images, (b) ISDNet, (c) WSDNet, (d) Groundtruth. Our method produces more accurate and detailed results, which are indicated by dotted boxes}
    \label{fig_vis}
\end{figure}

To show the effectiveness of WSDNet intuitively, we visualize and compare the results of several methods in Figure \ref{fig_vis}. 

\section{Conclusion}

The paper firstly proposes URUR, a large-scale dataset covering a wide range of scenes with full fine-grained dense annotations. It contains amounts of images with high enough resolution, a wide range of complex scenes, ultra-rich context and fine-grained annotations, which is far superior to all the existing UHR datasets. Furthermore, WSDNet is proposed to formulate a more efficient framework for UHR segmentation, where DWT-IWT paradigm is integrated to preserve more spatial details. Wavelet Smooth Loss (WSL) is designed to reconstruct original structured context and texture distribution. It is more concise, effective and stable than ordinary super-resolution loss. Extensive experiments show the remarkable superiority of URUR and WSDNet.

\section{Broader Impact}

Ultra-high image analysis has broadened the field of AI and Computer Vision researches, as well as poses extreme demands and challenges for models about both accuracy, inference speed and memory cost. Our work pushes the boundaries of Ultra-high image analysis. The URUR dataset build a new standard UHR benchmark for the community, which will benefit a wide range of natural disaster prevention, land resources utilization and urban construction planning applications.  The design of WSDNet can be generalized to the UHR ``Complicated Wild Scene Understanding". Even with these achievements, we realize that our work is not meant to be perfect, and there are still unpredictable challenges in the real world, depending on the specific application forms. In addition, our method can still help the research of natural scenes, from a more holistic and fine-grained perspective.

\section*{Acknowledgement}
This work was supported by the National Key R\&D Program of China under Grant 2020AAA0103902, the Anhui Provincial Natural Science Foundation under Grant 2108085UD12, the JKW Research Funds under Grant 20-163-14-LZ-001-004-01, NSFC (No. 62176155), Shanghai Municipal Science and Technology Major Project, China (2021SHZDZX0102). We acknowledge the support of GPU cluster built by MCC Lab of Information Science and Technology Institution, USTC.

{\small
\bibliographystyle{unsrt}
\bibliography{egbib}

\begin{thebibliography}{10}

\bibitem{pascal}
Mark Everingham, Luc~Van Gool, Christopher K.~I. Williams, John~M. Winn, and
  Andrew Zisserman.
\newblock The pascal visual object classes (voc) challenge., 2010.

\bibitem{coco}
Tsung-Yi Lin, Michael Maire, Serge Belongie, Lubomir Bourdev, Ross Girshick,
  James Hays, Pietro Perona, Deva Ramanan, C.~Lawrence Zitnick, and Piotr
  Dollár.
\newblock Microsoft coco: Common objects in context, 2014.

\bibitem{cityscapes}
Marius Cordts, Mohamed Omran, Sebastian Ramos, Timo Rehfeld, Markus Enzweiler,
  Rodrigo Benenson, Uwe Franke, Stefan Roth, and Bernt Schiele.
\newblock The cityscapes dataset for semantic urban scene understanding.
\newblock In {\em Proc. of the IEEE Conference on Computer Vision and Pattern
  Recognition (CVPR)}, 2016.

\bibitem{deepglobe}
I~Demir, K~Koperski, D~Lindenbaum, G~Pang, J~Huang, S~Basu, F~Hughes, D~Tuia,
  and R~Deepglobe Raskar.
\newblock Deepglobe 2018: A challenge to parse the earth through satellite
  images.
\newblock {\em CVPRW, pages 172–181}, 2018.

\bibitem{isic}
Philipp Tschandl, Cliff Rosendahl, and Harald Kittler.
\newblock The ham10000 dataset, a large collection of multi-source
  dermatoscopic images of common pigmented skin lesions.
\newblock {\em Scientific data}, 5(1):1--9, 2018.

\bibitem{udd}
Yu~Chen, Yao Wang, Peng Lu, Yisong Chen, and Guoping Wang.
\newblock Large-scale structure from motion with semantic constraints of aerial
  images.
\newblock In {\em Chinese Conference on Pattern Recognition and Computer Vision
  (PRCV)}, pages 347--359. Springer, 2018.

\bibitem{uavid}
Ye~Lyu, George Vosselman, Gui-Song Xia, Alper Yilmaz, and Michael~Ying Yang.
\newblock Uavid: A semantic segmentation dataset for uav imagery.
\newblock {\em ISPRS Journal of Photogrammetry and Remote Sensing}, 165:108 --
  119, 2020.

\bibitem{inria}
Emmanuel Maggiori, Yuliya Tarabalka, Guillaume Charpiat, and Pierre Alliez.
\newblock Can semantic labeling methods generalize to any city? the inria
  aerial image labeling benchmark.
\newblock In {\em 2017 IEEE International Geoscience and Remote Sensing
  Symposium (IGARSS)}, pages 3226--3229. IEEE, 2017.

\bibitem{ascher2007filmmaker}
Steven Ascher and Edward Pincus.
\newblock {\em The filmmaker's handbook: A comprehensive guide for the digital
  age}.
\newblock Penguin, 2007.

\bibitem{glnet}
Wuyang Chen, Ziyu Jiang, Zhangyang Wang, Kexin Cui, and Xiaoning Qian.
\newblock Collaborative global-local networks for memory-efficient segmentation
  of ultra-high resolution images.
\newblock In {\em Proceedings of the IEEE/CVF Conference on Computer Vision and
  Pattern Recognition}, pages 8924--8933, 2019.

\bibitem{fctl}
Qi~Li, Weixiang Yang, Wenxi Liu, Yuanlong Yu, and Shengfeng He.
\newblock From contexts to locality: Ultra-high resolution image segmentation
  via locality-aware contextual correlation.
\newblock In {\em Proceedings of the IEEE/CVF International Conference on
  Computer Vision}, pages 7252--7261, 2021.

\bibitem{isdnet}
Shaohua Guo, Liang Liu, Zhenye Gan, Yabiao Wang, Wuhao Zhang, Chengjie Wang,
  Guannan Jiang, Wei Zhang, Ran Yi, Lizhuang Ma, and Ke~Xu.
\newblock Isdnet: Integrating shallow and deep networks for efficient
  ultra-high resolution segmentation.
\newblock In {\em Proceedings of the IEEE/CVF Conference on Computer Vision and
  Pattern Recognition (CVPR)}, pages 4361--4370, June 2022.

\bibitem{goodfellow2016deep}
Ian Goodfellow, Yoshua Bengio, and Aaron Courville.
\newblock {\em Deep learning}.
\newblock MIT press, 2016.

\bibitem{zhang2015cross}
Cong Zhang, Hongsheng Li, Xiaogang Wang, and Xiaokang Yang.
\newblock Cross-scene crowd counting via deep convolutional neural networks.
\newblock In {\em Proceedings of the IEEE conference on computer vision and
  pattern recognition}, pages 833--841, 2015.

\bibitem{ji2019end}
Deyi Ji, Hongtao Lu, and Tongzhen Zhang.
\newblock End to end multi-scale convolutional neural network for crowd
  counting.
\newblock In {\em Eleventh international conference on machine vision (ICMV
  2018)}, volume 11041, pages 761--766. SPIE, 2019.

\bibitem{zhang2018context}
Hang Zhang, Kristin Dana, Jianping Shi, Zhongyue Zhang, Xiaogang Wang, Ambrish
  Tyagi, and Amit Agrawal.
\newblock Context encoding for semantic segmentation.
\newblock In {\em Proceedings of the IEEE conference on Computer Vision and
  Pattern Recognition}, pages 7151--7160, 2018.

\bibitem{feng2018challenges}
Weitao Feng, Deyi Ji, Yiru Wang, Shuorong Chang, Hansheng Ren, and Weihao Gan.
\newblock Challenges on large scale surveillance video analysis.
\newblock In {\em Proceedings of the IEEE conference on computer vision and
  pattern recognition workshops}, pages 69--76, 2018.

\bibitem{fcn}
Jonathan Long, Evan Shelhamer, and Trevor Darrell.
\newblock Fully convolutional networks for semantic segmentation.
\newblock In {\em Proceedings of the IEEE conference on computer vision and
  pattern recognition}, pages 3431--3440, 2015.

\bibitem{pspnet}
Hengshuang Zhao, Jianping Shi, Xiaojuan Qi, Xiaogang Wang, and Jiaya Jia.
\newblock Pyramid scene parsing network.
\newblock In {\em Proceedings of the IEEE conference on computer vision and
  pattern recognition}, pages 2881--2890, 2017.

\bibitem{deeplabv3+}
Liang-Chieh Chen, Yukun Zhu, George Papandreou, Florian Schroff, and Hartwig
  Adam.
\newblock Encoder-decoder with atrous separable convolution for semantic image
  segmentation.
\newblock In {\em Proceedings of the European conference on computer vision
  (ECCV)}, pages 801--818, 2018.

\bibitem{ocrnet}
Yuhui Yuan, Xilin Chen, and Jingdong Wang.
\newblock Object-contextual representations for semantic segmentation.
\newblock In {\em European conference on computer vision}, pages 173--190.
  Springer, 2020.

\bibitem{segformer}
Enze Xie, Wenhai Wang, Zhiding Yu, Anima Anandkumar, Jose~M Alvarez, and Ping
  Luo.
\newblock Segformer: Simple and efficient design for semantic segmentation with
  transformers.
\newblock {\em NeurIPS}, 2021.

\bibitem{setr}
Sixiao Zheng, Jiachen Lu, Hengshuang Zhao, Xiatian Zhu, Zekun Luo, Yabiao Wang,
  Yanwei Fu, Jianfeng Feng, Tao Xiang, Philip~H.S. Torr, and Li~Zhang.
\newblock Rethinking semantic segmentation from a sequence-to-sequence
  perspective with transformers.
\newblock In {\em CVPR}, 2021.

\bibitem{stlnet}
Lanyun Zhu, Deyi Ji, Shiping Zhu, Weihao Gan, Wei Wu, and Junjie Yan.
\newblock Learning statistical texture for semantic segmentation.
\newblock In {\em Proceedings of the IEEE/CVF Conference on Computer Vision and
  Pattern Recognition (CVPR)}, pages 12537--12546, June 2021.

\bibitem{cdgc}
Hanzhe Hu, Deyi Ji, Weihao Gan, Shuai Bai, Wei Wu, and Junjie Yan.
\newblock Class-wise dynamic graph convolution for semantic segmentation.
\newblock In {\em European Conference on Computer Vision}, pages 1--17.
  Springer, 2020.

\bibitem{zhou2020graph}
Jie Zhou, Ganqu Cui, Shengding Hu, Zhengyan Zhang, Cheng Yang, Zhiyuan Liu,
  Lifeng Wang, Changcheng Li, and Maosong Sun.
\newblock Graph neural networks: A review of methods and applications.
\newblock {\em AI open}, 1:57--81, 2020.

\bibitem{cagcn}
Deyi Ji, Haoran Wang, Hanzhe Hu, Weihao Gan, Wei Wu, and Junjie Yan.
\newblock Context-aware graph convolution network for target re-identification.
\newblock {\em arXiv preprint arXiv:2012.04298}, 2020.

\bibitem{wu2019simplifying}
Felix Wu, Amauri Souza, Tianyi Zhang, Christopher Fifty, Tao Yu, and Kilian
  Weinberger.
\newblock Simplifying graph convolutional networks.
\newblock In {\em International conference on machine learning}, pages
  6861--6871. PMLR, 2019.

\bibitem{ipgn}
Haoran Wang, Licheng Jiao, Fang Liu, Lingling Li, Xu~Liu, Deyi Ji, and Weihao
  Gan.
\newblock Ipgn: Interactiveness proposal graph network for human-object
  interaction detection.
\newblock {\em IEEE Transactions on Image Processing}, 30:6583--6593, 2021.

\bibitem{icnet}
Hengshuang Zhao, Xiaojuan Qi, Xiaoyong Shen, Jianping Shi, and Jiaya Jia.
\newblock Icnet for real-time semantic segmentation on high-resolution images.
\newblock In {\em Proceedings of the European conference on computer vision
  (ECCV)}, pages 405--420, 2018.

\bibitem{bisenetv2}
Changqian Yu, Changxin Gao, Jingbo Wang, Gang Yu, Chunhua Shen, and Nong Sang.
\newblock Bisenet v2: Bilateral network with guided aggregation for real-time
  semantic segmentation.
\newblock {\em International Journal of Computer Vision}, 129(11):3051--3068,
  2021.

\bibitem{stdc}
Mingyuan Fan, Shenqi Lai, Junshi Huang, Xiaoming Wei, Zhenhua Chai, Junfeng
  Luo, and Xiaolin Wei.
\newblock Rethinking bisenet for real-time semantic segmentation.
\newblock In {\em Proceedings of the IEEE/CVF conference on computer vision and
  pattern recognition}, pages 9716--9725, 2021.

\bibitem{skd}
Yifan Liu, Ke~Chen, Chris Liu, Zengchang Qin, Zhenbo Luo, and Jingdong Wang.
\newblock Structured knowledge distillation for semantic segmentation.
\newblock In {\em Proceedings of the IEEE/CVF Conference on Computer Vision and
  Pattern Recognition}, pages 2604--2613, 2019.

\bibitem{sstkd}
Deyi Ji, Haoran Wang, Mingyuan Tao, Jianqiang Huang, Xian-Sheng Hua, and
  Hongtao Lu.
\newblock Structural and statistical texture knowledge distillation for
  semantic segmentation.
\newblock In {\em Proceedings of the IEEE/CVF Conference on Computer Vision and
  Pattern Recognition}, pages 16876--16885, 2022.

\bibitem{cascadepsp}
Ho~Kei Cheng, Jihoon Chung, Yu-Wing Tai, and Chi-Keung Tang.
\newblock Cascadepsp: Toward class-agnostic and very high-resolution
  segmentation via global and local refinement.
\newblock In {\em Proceedings of the IEEE/CVF Conference on Computer Vision and
  Pattern Recognition}, pages 8890--8899, 2020.

\bibitem{gpwformer}
Deyi Ji, Feng Zhao, and Hongtao Lu.
\newblock Guided patch-grouping wavelet transformer with spatial congruence for
  ultra-high resolution segmentation.
\newblock {\em International Joint Conference on Artificial Intelligence},
  2023.

\bibitem{erm}
Gellért Máttyus, Shenlong Wang, Sanja Fidler, and Raquel Urtasun.
\newblock Enhancing road maps by parsing aerial images around the world.
\newblock In {\em 2015 IEEE International Conference on Computer Vision
  (ICCV)}, pages 1689--1697, 2015.

\bibitem{wang2021learning}
Haoran Wang, Licheng Jiao, Fang Liu, Lingling Li, Xu~Liu, Deyi Ji, and Weihao
  Gan.
\newblock Learning social spatio-temporal relation graph in the wild and a
  video benchmark.
\newblock {\em IEEE Transactions on Neural Networks and Learning Systems},
  2021.

\bibitem{wavelet_cnn}
Pengju Liu, Hongzhi Zhang, Kai Zhang, Liang Lin, and Wangmeng Zuo.
\newblock Multi-level wavelet-cnn for image restoration.
\newblock {\em CoRR}, abs/1805.07071, 2018.

\bibitem{wavevit}
Ting Yao, Yingwei Pan, Yehao Li, Chong-Wah Ngo, and Tao Mei.
\newblock Wave-vit: Unifying wavelet and transformers for visual representation
  learning.
\newblock In {\em Proceedings of the European conference on computer vision
  (ECCV)}, 2022.

\bibitem{mmseg2020}
MMSegmentation Contributors.
\newblock {MMSegmentation}: Openmmlab semantic segmentation toolbox and
  benchmark.
\newblock \url{https://github.com/open-mmlab/mmsegmentation}, 2020.

\bibitem{unet}
Olaf Ronneberger, Philipp Fischer, and Thomas Brox.
\newblock U-net: Convolutional networks for biomedical image segmentation.
\newblock In {\em International Conference on Medical image computing and
  computer-assisted intervention}, pages 234--241. Springer, 2015.

\bibitem{bisenet}
Changqian Yu, Jingbo Wang, Chao Peng, Changxin Gao, Gang Yu, and Nong Sang.
\newblock Bisenet: Bilateral segmentation network for real-time semantic
  segmentation.
\newblock In {\em Proceedings of the European conference on computer vision
  (ECCV)}, pages 325--341, 2018.

\bibitem{danet}
Jun Fu, Jing Liu, Haijie Tian, Yong Li, Yongjun Bao, Zhiwei Fang, and Hanqing
  Lu.
\newblock Dual attention network for scene segmentation.
\newblock In {\em Proceedings of the IEEE/CVF conference on computer vision and
  pattern recognition}, pages 3146--3154, 2019.

\bibitem{ppn}
Tong Wu, Zhenzhen Lei, Bingqian Lin, Cuihua Li, Yanyun Qu, and Yuan Xie.
\newblock Patch proposal network for fast semantic segmentation of
  high-resolution images.
\newblock In {\em Proceedings of the AAAI Conference on Artificial
  Intelligence}, 2020.

\bibitem{pointrend}
Alexander Kirillov, Yuxin Wu, Kaiming He, and Ross Girshick.
\newblock Pointrend: Image segmentation as rendering.
\newblock In {\em Proceedings of the IEEE/CVF conference on computer vision and
  pattern recognition}, pages 9799--9808, 2020.

\bibitem{magnet}
Chuong Huynh, Anh~Tuan Tran, Khoa Luu, and Minh Hoai.
\newblock Progressive semantic segmentation.
\newblock In {\em Proceedings of the IEEE/CVF Conference on Computer Vision and
  Pattern Recognition}, pages 16755--16764, 2021.

\bibitem{deformable}
Jifeng Dai, Haozhi Qi, Yuwen Xiong, Yi~Li, Guodong Zhang, Han Hu, and Yichen
  Wei.
\newblock Deformable convolutional networks.
\newblock In {\em Proceedings of the IEEE international conference on computer
  vision}, pages 764--773, 2017.

\end{thebibliography}
}

\end{document}